# Optimization of Inter-Subnet Belief Updating in Multiply Sectioned Bayesian Networks

**Y. Xiang**
Department of Computer Science, University of Regina
Regina, Saskatchewan, Canada S4S 0A2, yxiang@cs.uregina.ca

## Abstract

Recent developments show that Multiply Sectioned Bayesian Networks (MSBNs) can be used for diagnosis of natural systems as well as for model-based diagnosis of artificial systems. They can be applied to single-agent oriented reasoning systems as well as multi-agent distributed reasoning systems.

Belief propagation between a pair of subnets plays a central role in maintenance of global consistency in a MSBN. This paper studies the operation `UpdateBelief`, presented originally with MSBNs, for inter-subnet propagation. We analyze how the operation achieves its intended functionality, which provides hints for improving its efficiency.

New versions of `UpdateBelief` are then defined that reduce the computation time for inter-subnet propagation. One of them is optimal in the sense that the minimum amount of computation for coordinating multi-linkage belief propagation is required. The optimization problem is solved through the solution of a graph-theoretic problem: the minimum weight open tour in a tree.

## 1 Introduction

Multiply sectioned Bayesian networks (MSBNs) [9] is an extension of Bayesian networks (BNs) [4, 3, 1], originally developed for modular knowledge representation and more efficient inference computation in large application domains [7]. The basic assumption of MSBNs is *localization* [9, 8, 7]: Subdomains of the target domain are loosely coupled such that evidence and queries focus on one subdomain for a period of time before shifting to a different subdomain. Based on localization, a MSBN represents a large domain by a set of interrelated Bayesian subnets, such that inference computation can be confined within one subnet at a time.

Two recent developments in probabilistic reasoning using BNs widened the scope of potential applicability of the MSBN representation/inference formalism.

Srinivas [5] proposed a hierarchical approach for model-based diagnoses. The representation formalism used can be viewed as a special case of MSBNs. For example, the set of input nodes $I$, output node $O$, mode node $M$, and dummy node $D$ [5], which forms an interface between a higher level and a lower level in the hierarchy, is a d-sepset [9]. The 'composite joint tree' [5] corresponds to the 'hypertree' [9]. The way in which inference is performed in the composite join tree corresponds to the operation *ShiftAttention* [9]. Therefore, his work showed that MSBNs can be applied to diagnosis of natural systems (e.g., human body [7]) as well as artificial systems (e.g., electronic circuits [5]).

Instead of viewing a MSBN as representing multiple perspectives of a single reasoning agent, a MSBN can be viewed as representing multiple agents in a domain each of which holds one perspective of the domain. Following this semantics, Xiang [6] extended MSBNs to distributed multi-agent probabilistic reasoning, where multiple agents (subnets) collect local evidence asynchronously *in parallel* and exchange information *infrequently* to achieve a common goal. We shall sometime use the terms 'subnet' and 'agent' interchangeably in the paper.

Given the widened applicability of the MSBN formalism, this paper reexamines the key inference operation `UpdateBelief` [9] of MSBNs for propagating belief from one subnet to another. We propose two new versions of `UpdateBelief` to improve its efficiency. We compare the two improvements and indicate their trade-offs.

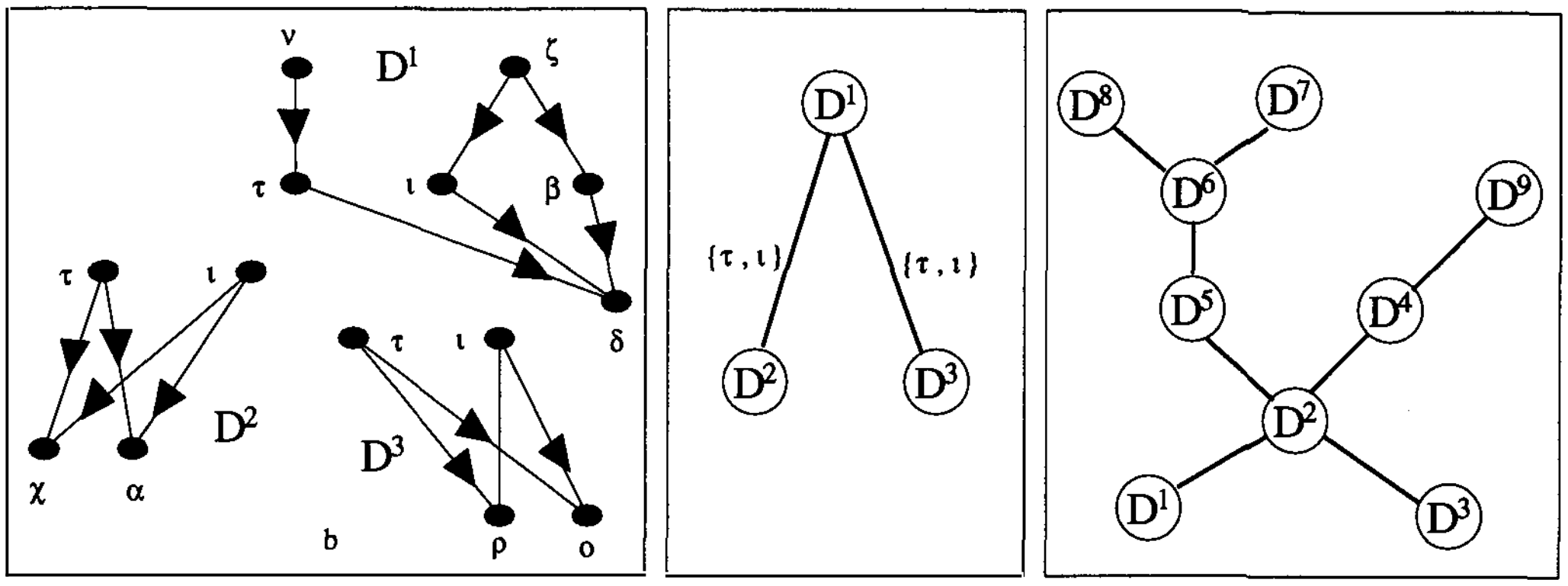

Figure 1: Left: A three-subnet MSBN. Middle: The hypertree organization of the MSBN in the left. Right: A general hypertree structured MSBN.

Section 2 briefly reviews the MSBN formalism and defines the concepts that the rest of the paper depends on. Section 3 describes the role of **UpdateBelief** intuitively, and analyzes how it achieves its intended functionality by proving a theorem. Based on the hints provided by the proof, Section 4 proposes a new version of **UpdateBelief** that reduces its computation time. Section 5 discusses intuitively the possibility of further efficiency improvement. The issue is then formulated as an optimization problem in tree traversals with the solution given in Section 6. We summerize the difference of the three versions of **UpdateBelief** and their trade-offs in Section 7.

## 2 Overview of MSBNs

This section briefly reviews the MSBN formalism. We shall use freely the formal results in [9, 6], subject to some simplification, for concepts that the rest of the paper depends on.

A MSBN $M$ consists of a set of interrelated Bayesian subnets. Each subnet shares a non-empty set $I$ of variables with at least one other subnet. This interfacing set $I$ must be a *d-sepset*, which ensures that, when the pair of subnets is isolated from the rest of $M$, $I$ renders the two subnets conditionally independent. Figure 1 (left) shows a three-subnet MSBN.

Subnets of $M$ are organized into a *hypertree* structure. The hypernodes are subnets of $M$. The hyperlinks are d-sepsets between subnets. A hypertree structured $M$ ensures that each hyperlink render the two parts of $M$ that it connects conditionally independent. Figure 1 (middle) shows the hypertree organization of the MSBN in Figure 1 (left). Figure 1 (right) depicts a general hypertree structured MSBN.

The hypertree structured $M$ is converted into a *linked junction forest* (LJF) $F$ of the identical structure as its run time representation. Each hypernode in the hypertree is a *junction tree* (JT) (clique tree) converted from the corresponding subnet in the hypertree structured $M$. Evidence can be propagated between JTs of $F$ by passing the probability distribution (PD) of $I$, which would not be efficient if the cardinality of $I$ is large. The efficiency can be improved by exploiting the following structure internal to $I$.

**Definition 1 (host tree)** *Let $I$ be the d-sepset between JTs $T^a$ and $T^b$. A* **host tree** *$H$ of $T^a$ relative to $T^b$ is obtained by recursively removing every leaf clique $C$ of $T^a$ that satisfies one of the following conditions. (1) $C \cap I = \phi$. (2) $C \cap I$ is a subset of another clique on the current $H$.*

$H$ is the minimum subtree that contains $I$. Only cliques in $H$ are involved *directly* with the evidence propagation between JTs. $H$ is further reduced to a linkage tree for definition of propagation channels.

**Definition 2 (linkage tree)** *Let $I$ be the d-sepset between JTs $T^a$ and $T^b$. A* **linkage tree** *of $T^a$ relative to $T^b$ is obtained by recursively removing nodes (variables contained in a clique) or cliques from the host tree $H$ of $T^a$ as follows.*
*(1) If a node $x \notin I$ is contained in a single clique $C$, remove $x$ from $C$.*
*(2) If a clique $C$ becomes a subset of an adjacent clique $D$ after the above operation, union $C$ into $D$.*

The removal of $x$ corresponds to a marginalization operation on $x$. The union of $C$ into $D$ deletes $C$ and the link $(C, D)$ from $H$, and reconnects to $D$ all cliques originally adjacent to $C$. *Correct* evidence propagation between JTs of $F$ can be achieved iff every linkage tree contains exactly the nodes in the corresponding $I$.

Each clique in a linkage tree $L$ is a *linkage*. Each linkage in $L$ has a corresponding *host* clique $C$ in $H$ and hence the name host tree. The linkages in $L$ are indexed as $L_1, L_2, \ldots$ in any order consistent with $L$. That is, for every $i$ there is a unique $j < i$ such that $L_j$ is adjacent to $L_i$.

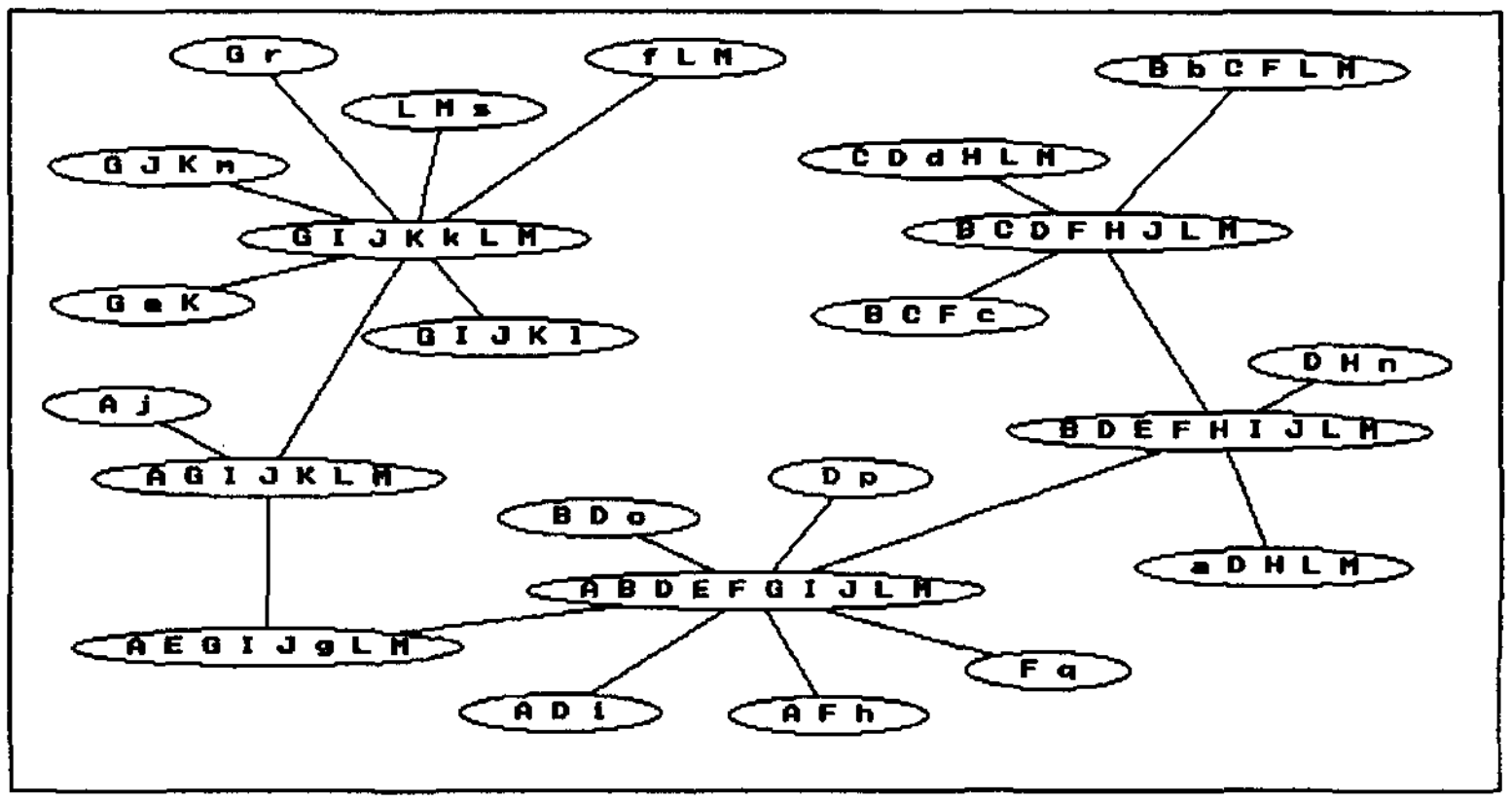

Figure 2: A JT where each upper case letter represents a variable in the d-sepset, and each lower case letter represents a variable not in the d-sepset.

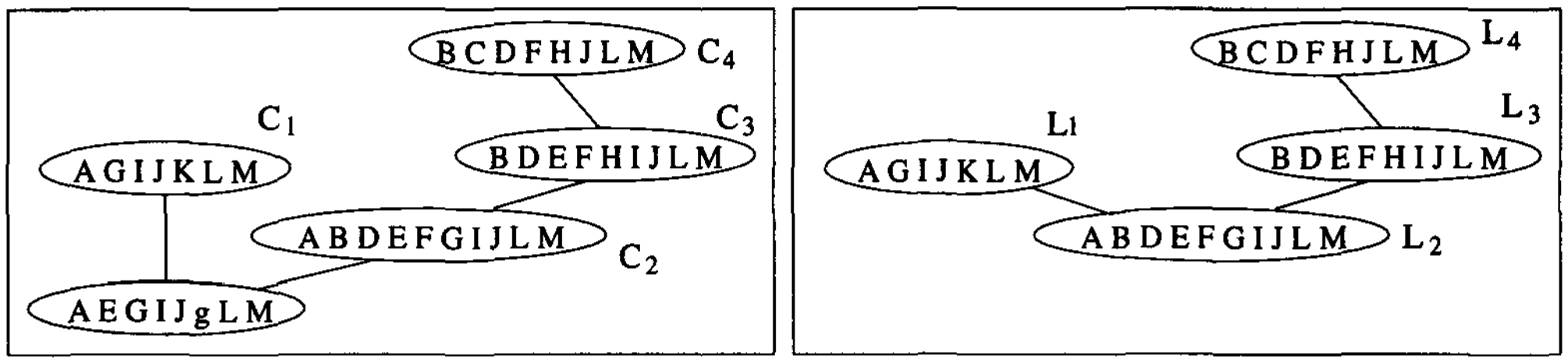

Figure 3: The host tree (left) and linkage tree (right) of the JT in Figure 2.

The tree structure of $L$ allows the PD on $I$ (in the form of belief table $B(I)$) be propagated between JTs by passing only belief tables on individual linkages since $B(I) = \prod B(L_j)/\prod B(R_k)$ (see [9] for definitions of operations on belief tables) where $B(L_j) = \sum_{U_j \setminus L_j} B(U_j)$ is the belief table of the linkage $L_j$ whose host is $U_j$, and $B(R_k)$ is the belief table of a sepset (intersection of two adjacent cliques) in $L$. This may reduce the propagation traffic significantly when $L$ consists of many small cliques.

The conversion of $M$ into $F$ ensures that the joint probability distribution (JPD) of $F$ assembled from belief tables of $F$ be equal to that of $M$.

Figure 2 shows a JT whose d-sepset with an adjacent JT is $I = \{A, B, C, D, E, F, G, H, I, J, K, L, M\}$. The host tree of the JT is shown in Figure 3 (left). The linkage tree is shown in Figure 3 (right). The clique at the lower left corner of Figure 3 (left) has been deleted since after the variable $g$ in it is removed, it becomes a subset of clique $C_2$ and itself is unioned into $C_2$. The indexing of linkages are shown in Figure 3 (right) and, for each linkage, its host is labeled in Figure 3 (left) with the same index. Each linkage happens to be identical to its host in this example. But in general this is not the case.

To answer queries by efficient local computation in $F$, it must be made consistent. $F$ is *locally consistent* if all JTs are *internally consistent*, i.e., when marginalized onto the same set of variables, different belief tables in a JT yield the same PD. $F$ is *boundary consistent* if each pair of adjacent JTs are consistent with respect to their d-sepset $I$. $F$ is *globally consistent* iff it is both locally consistent and boundary consistent.

A set of operations [9, 6] are developed to achieve consistency in $F$ during evidential reasoning. `BeliefInitialization` establishes initial global consistency. `DistributeEvidence` causes an outward belief propagation within a JT, and brings the JT internally consistent after evidence on variables in a single clique has been entered. `UnifyBelief` brings a JT internally consistent after evidence on variables in multiple cliques has been entered. `EnterEvidence` updates belief in a JT in light of new evidence, and brings the JT internally consistent again by calling either `DistributeEvidence` or `UnifyBelief`. `AbsorbThroughLinkage` brings two linkage hosts in different JTs into consistency. `UpdateBelief` updates the belief of a JT $T$ relative to an adjacent JT, and brings $T$ internally consistent. In a single agent MSBN, `ShiftAttention` allows the user to enter multiple pieces of evidence into a JT of current attention, and, when the user shifts attention to a target JT, maintains consistency along the hyperpath in the hypertree structured $F$ from the current JT to the target. In a multi-agent MSBN, `CommunicateBelief` regains global consistency after multiple agents have obtained evidence asynchronously in parallel. Both `ShiftAttention` and `CommunicateBelief` rely on `UpdateBelief` for inter-subnet belief propagation.

## 3 Insight into the UpdateBelief Operation

The operation `UpdateBelief` plays a central role in inter-subnet communication. It is defined as follows.

**Operation 3 (UpdateBelief [9])** *Let $\{L_1, \ldots, L_m\}$ be the set of linkages of a JT $T^a$ relative to $T^b$. Let $U_i^a$ and $U_i^b$ be the linkage hosts of $L_i$ ($i = 1, \ldots, m$) in $T^a$ and $T^b$, respectively. When* `UpdateBelief` *is initiated by $T^a$ relative to $T^b$, the following is performed. For each $i$ (in ascending order)* `AbsorbThroughLinkage` *is called in $U_i^a$ to absorb from $U_i^b$ through $L_i$, followed by a* `DistributeEvidence` *called in $U_i^a$.*

Intuitively, `UpdateBelief` can be understood as follows. In order to bring two adjacent JTs into consistency, we need to propagate the belief table $B(I)$ from $T^b$ to $T^a$. Since the size of $B(I)$ is exponential to the size of $I$, the propagation can be expensive. In multi-agent MSBNs with remotely located agents, we need to pass $B(I)$ through communication channels. `UpdateBelief` makes use of conditional independence among members of the d-sepset, and propagates $B(I)$ by only passing the belief tables $B(L_i)$, which are collectively smaller in size when $B(I)$ is large. For example, if the d-sepset $I$ contains ten binary variables, $B(I)$ has a size of 1024. If $I = L_1 \cup L_2 \cup L_3$ with each $L_i$ containing 5 variables, then the three belief tables on linkages have a total size 96.

However, this savings in propagation traffic has a price to pay. The belief propagation over multiple linkages must be coordinated to achieve the intended effect. As analyzed in [9], each `AbsorbThroughLinkage` should be followed by the operation `DistributeEvidence` in $T^a$. When $T^a$ is large, `DistributeEvidence` performed repeatedly can be expensive. Based on the argument "communication is slower than computation" [2], the tradeoff is justified [6] by observing that `DistributeEvidence` involves only local computation within a subnet. Admitting that communication savings should be preferred over computation savings, this paper aims to improve the efficiency of the local computation as much as possible.

In the original presentation of `UpdateBelief` [9], how the operation works is not fully analyzed. The proof of the following theorem gains further insight into this operation, and provides hints for improvement of its efficiency.

**Theorem 4** *Let $I$ be the d-sepset between JTs $T^a$ and $T^b$. Let $\{L_1, \ldots, L_m\}$ be the set of linkages. Let the two JTs be internally consistent. Let $B(I^a)$ ($B(I^b)$) be the belief table on $I$ defined by marginalization of $B(T^a)$ ($B(T^b)$). After* `UpdateBelief`*, $B'(T^a) = B(T^a) * B(I^b)/B(I^a)$, and $T^a$ is internally consistent.*

Proof: We prove by induction on the index of linkages. `AbsorbThroughLinkage` in `UpdateBelief` is performed by $T^a$ in the order $L_1, \ldots, L_m$. After the performance of `AbsorbThroughLinkage` through $L_1$, we have $B_1(T^a) = B(T^a) * B(L_1^b)/B(L_1^a)$.

After `AbsorbThroughLinkage` is performed through $L_2$, we have $B_2(T^a) = B_1(T^a) * B(L_2^b)/B_1(L_2^a)$. Note that the two $B()$ in the right-hand side of the previous equation are now replaced by $B_1()$. The appearance of $B_1(L_2^a)$ instead of $B(L_2^a)$ is due to the first `DistributeEvidence` that follows the first `AbsorbThroughLinkage`. After `DistributeEvidence`, $T^a$ is internally consistent and $B(L_2^a)$ is updated into $B_1(L_2^a)$. We also obtain the following equation:

$$B_1(L_2^a) = B(L_2^a) * B(L_1 \cap^b L_2)/B(L_1 \cap^a L_2),$$

where '$\cap^a$' signifies that the intersection is defined in $T^a$ and so is the $B()$.

Substituting $B_1(T^a)$ and $B_1(L_2^a)$, we obtain

$$\begin{aligned} B_2(T^a) &= B(T^a) * \frac{B(L_1^b) * B(L_2^b)/B(L_1 \cap^b L_2)}{B(L_1^a) * B(L_2^a)/B(L_1 \cap^a L_2)} \\ &= B(T^a) * \frac{B(L_1 \cup^b L_2)}{B(L_1 \cup^a L_2)}, \end{aligned}$$

where the second equality holds because of the way in which linkages are defined and indexed (Section 2).

Assume that, after `AbsorbThroughLinkage` is performed through $L_i$ followed by `DistributeEvidence`, we have the following two equations.

$$B_i(T^a) = B(T^a) * \frac{B(\bigcup_{j\le i}^b L_j)}{B(\bigcup_{k\le i}^a L_k)} \quad (1)$$

$$B_i(L_{i+1}^a) = B(L_{i+1}^a) * \frac{B((\bigcup_{j\le i}^b L_j) \cap^b L_{i+1})}{B((\bigcup_{k\le i}^a L_k) \cap^a L_{i+1})} \quad (2)$$

After `AbsorbThroughLinkage` is performed through $L_{i+1}$, we have $B_{i+1}(T^a) = B_i(T^a) * B(L_{i+1}^b)/B_i(L_{i+1}^a)$.

By substituting $B_i(T^a)$ and $B_i(L_{i+1}^a)$, it yields

$$\begin{aligned} &B_{i+1}(T^a) \\ &= B(T^a) * \frac{B(\bigcup_{j\le i}^b L_j) * B(L_{i+1}^b)/B((\bigcup_{j\le i}^b L_j) \cap^b L_{i+1})}{B(\bigcup_{k\le i}^a L_k) * B(L_{i+1}^a)/B((\bigcup_{k\le i}^a L_k) \cap^a L_{i+1})} \\ &= B(T^a) * \frac{B(\bigcup_{j\le i+1}^b L_j))}{B(\bigcup_{k\le i+1}^a L_k))}, \quad (3) \end{aligned}$$

where the second equality holds because of the way in which linkages are defined and indexed (Section 2).

After the DistributeEvidence is performed, $T^a$ is internally consistent, and it follows that

$$B_{i+1}(L^a_{i+2}) = B(L^a_{i+2}) * \frac{B((\bigcup^b_{j\leq i+1} L_j) \cap^b L_{i+2})}{B((\bigcup^a_{k\leq i+1} L_k) \cap^a L_{i+2})}. \quad (4)$$

From the inductive assumptions (1) and (2), we have now proven the conditions (3) and (4). Therefore, after AbsorbThroughLinkage is performed through the last linkage $L_m$, we obtain the updated belief

$$\begin{aligned} B'(T^a) &= B_m(T^a) = B(T^a) * \frac{B(\bigcup^b_{j\leq m} L_j))}{B(\bigcup^a_{k\leq m} L_k))} \\ &= B(T^a) * B(I^b)/B(I^a), \end{aligned}$$

where the last equality holds due to the way in which the linkage tree is defined. $T^a$ is internally consistent after the last DistributeEvidence. □

## 4 Efficiency Improvement of UpdateBelief

In the proof of Theorem 4, it is observed that the inductive conclusion on $B_{i+1}(L^a_{i+2})$ (equation (4)) can be proven as long as the *host tree* (Section 2) is made consistent after the AbsorbThroughLinkage through $L_{i+1}$. The consistency of the *entire* JT is not necessary. Hence belief propagation beyond the boundary of the host tree, as performed by DistributeEvidence, is not necessary. We therefore define a new operation DistributeEvidenceOnHostTree which is the same as DistributeEvidence except that it terminates at the leaves of the host tree. For example, if DistributeEvidenceOnHostTree is called in the clique $\{A, B, D, E, F, G, I, J, L, M\}$ in the JT of Figure 2 (labeled $C_2$ in Figure 3 (left)), the belief propagation will proceed along two chains only: One from $C_2$ to $C_1$ and the other from $C_2$ to $C_4$.

Replacing DistributeEvidence by the new operation, we can define a new version of UpdateBelief.

**Operation 5 (UpdateBelief2)**
*Let $\{L_1, \ldots, L_m\}$ be the set of linkages of a JT $T^a$ relative to $T^b$. Let $U^a_i$ and $U^b_i$ be the linkage hosts of $L_i$ ($i = 1, \ldots, m$) in $T^a$ and $T^b$, respectively. When* UpdateBelief2 *is initiated by $T^a$ relative to $T^b$, the following is performed.*

*For $i = 1$ through $m$,* AbsorbThroughLinkage *is called in $U^a_i$ to absorb from $U^b_i$ through $L_i$. For $i = 1, \ldots, m-1$, it is followed by* DistributeEvidenceOnHostTree *called in $U^a_i$. For $i = m$, it is followed by* DistributeEvidence *called in $U^a_m$.*

**Corollary 6** *Let $I$ be the d-sepset between JTs $T^a$ and $T^b$. Let $\{L_1, \ldots, L_m\}$ be the set of linkages. Let the two JTs be internally consistent. Let $B(I^a)$ ($B(I^b)$) be the belief table on $I$ defined by marginalization of $B(T^a)$ ($B(T^b)$). After* UpdateBelief2, *$B'(T^a) = B(T^a) * B(I^b)/B(I^a)$, and $T^a$ is internally consistent.*

Proof: After each DistributeEvidenceOnHostTree, the host tree is internally consistent. Hence, all intermediate results on $B_i(T^a)$ and $B_i(L^a_{i+1})$ in the proof of Theorem 4 are still valid except that $T^a$ as a whole is not internally consistent until after the performance of DistributeEvidence at the end of UpdateBelief2.

□

UpdateBelief2 performs repeatedly DistributeEvidenceOnHostTree instead of DistributeEvidence. Computational savings are obtained by not having to propagate belief beyond the host tree for a number of times proportional to the number of linkages. When the host tree is significantly smaller than the JT, the savings can also be significant.

## 5 Further Improvement of UpdateBelief

Examination of the proof of Theorem 4 shows that propagation of belief to the entire host tree is still unnecessary. For the result of the theorem to be valid, it is sufficient to update $B_{i-1}(L^a_{i+1})$ to $B_i(L^a_{i+1})$, after AbsorbThroughLinkage through $L_i$. This implies that, between two successive AbsorbThroughLinkage, it is sufficient to propagate the new belief only to the next host clique. Based on this idea, a more efficient UpdateBelief can be defined. We illustrate the new operation with an example.

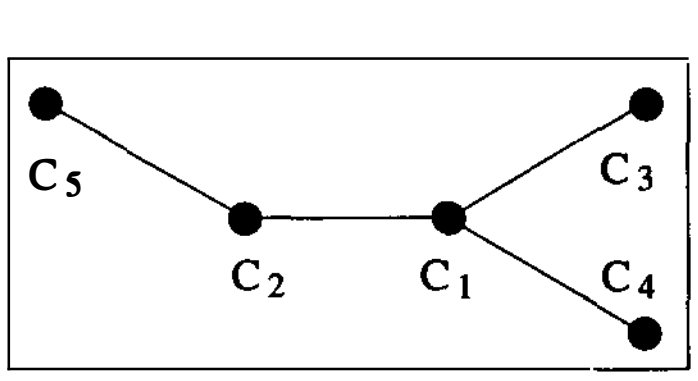

Figure 4: A host tree with five linkage hosts.

Consider the host tree in Figure 4. We assume that each clique is a host, and each clique is indexed by the index of the corresponding linkage. Suppose AbsorbThroughLinkage is performed in the order $i = 1, \ldots, 5$. After AbsorbThroughLinkage through $L_1$, we propagate the new belief in $C_1$ to $C_2$ (one inter-clique propagation). After AbsorbThroughLinkage through $L_2$, we propagate the new belief in $C_2$ to $C_1$ and then to $C_3$ (two inter-clique propagations). Propagating new belief this fashion, we need to perform $1+2+2+3 = 8$ inter-clique belief propagations, before the AbsorbThroughLinkage through $L_5$. Compared to $4+4+4+4=16$ inter-clique propagations needed in `UpdateBelief2`, there is an about 50% computational savings.

Additional savings can be obtained by optimizing the new operation. We note that **`AbsorbThroughLinkage`** is performed in the ascending order of linkage indexes. However, linkages can be indexed by any order consistent with the host tree (Section 2). We therefore have the freedom to choose the order that can maximize the computational savings.

If we choose the order $i = 5, 2, 1, 3, 4$ for the host tree in Figure 4, we only need to perform $1+1+1+2=5$ inter-clique belief propagations: Three propagations less than the previous order.

In the next section, we present the result on how to determine the optimal order of linkages (the order for performing **`AbsorbThroughLinkage`**) given a host tree.

## 6 Optimization of UpdateBelief

The problem of finding the optimal order for the performance of the operation **`AbsorbThroughLinkage`** in **`UpdateBelief`** can be abstracted as follows.

**Problem Statement 7** *Given a weighted tree of n nodes, number the nodes from 1 to n such that* $\sum_{i=1}^{n-1} w(i, i+1)$ *is minimized, where* $w(i, i+1)$ *is the path weight from node i to node* $i+1$ *according to the numbering.*

The tree in this model corresponds to the given host tree. Each node corresponds to a host clique. The link weight corresponds to the amount of computation for propagating belief from one clique to an adjacent clique in the host tree. The model assumes that every node is a host. If this is not the case, the propagation through non-host nodes can be modeled into the link weights such that the resultant tree has no non-host node.

We define the concept *tour* to be used in solving the above problem.

**Definition 8 (tour)** *A* **`tour`** *of a graph is a walk that visits each node at least once. A* **`closed tour`** *is a tour that starts and ends with the identical node. Otherwise, it is an* **`open tour`**.

The problem can now be equivalently expressed as follows.

**Problem Statement 9** *Given a weighted tree of n nodes, find an open tour with the minimum weight.*

In order to develop an algorithm that solves the above problem, we first study a closed tour, since, as will be shown, (1) the problem of a minimum weight closed tour can be solved easily; and (2) the minimum weight open tour has a simple relationship with the minimum weight closed tour. To simplify the intermediate derivations, we assume that all link *weights* are identical. We then simply deal with a minimum *length* tour with the length of each link being one. We remove this assumption at the end.

**Lemma 10** *A closed tour of a tree with the minimum length traverses each link exactly twice.*

Proof: We prove by induction. The statement is trivially true for a tree with only one link. Assume that it is true for a tree with $k$ link(s).

Consider a tree with $k+1$ links. For any leaf $x$ and its adjacent node $y$, if we remove $x$ and the link $(x, y)$, the resultant subgraph is a tree with $k$ link(s). By assumption, it has a minimum length closed tour that traverses each link exactly twice. To include $x$ and $(x, y)$ in the closed tour, one must at least travel from $y$ to $x$ and then come back. This completes a closed tour of the original tree with the minimum additional link traversal. □

For example, a closed tour of the minimum length for the tree in Figure 5 is $(C_5, C_2, C_6, C_7, C_8, C_7, C_6, C_2, C_1, C_9, C_{10}, C_9, C_1, C_3, C_1, C_4, C_1, C_2, C_5)$. The length of the tour is 18, which is twice of the number of links of the tree.

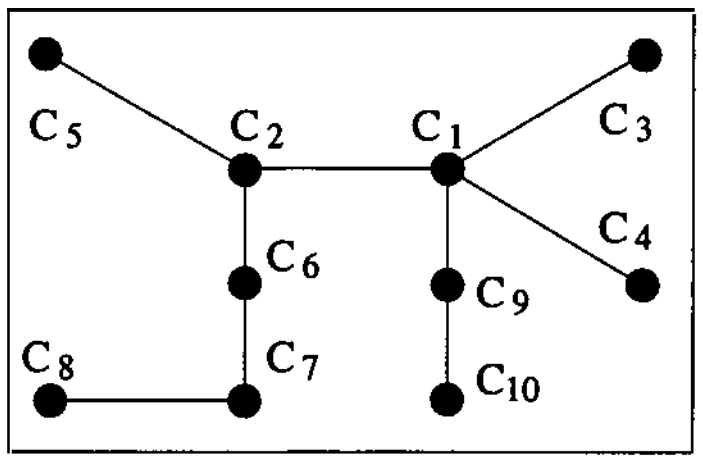

Figure 5: A tree of nine links.

We define a *terminal chain* to be used in the following discussion.

**Definition 11** *A* **`terminal chain`** *in a tree is a path (a walk without repeated nodes) that terminates at both ends by leaf nodes.*

For instance, one terminal chain in Figure 5 is $(C_5, C_2, C_6, C_7, C_8)$. The simple open path $(C_{10}, C_9, C_1)$ is not a terminal chain, since $C_1$ is not a leaf.

We now establish the relation between a minimum length closed tour and a minimum length open tour through a terminal chain.

**Lemma 12** *An open tour of a tree can be constructed such that its length is the length of a minimum length*

*closed tour minus that of a terminal chain.*

Proof: It suffices to show that an open tour can be constructed by reconnecting a minimum length closed tour $r$ such that a terminal chain is traversed only once.

Let $l$ be an arbitrary terminal chain in the tree. Start from one end of $l$ and travel along the chain. For each internal node $y$ in $l$, let the two adjacent nodes of $y$ on $l$ be $x$ and $z$, and let the direction of the traversal be from $x$ to $z$. At each $y$ of degree 3 or more, traverse first an adjacent node $u$ ($u \neq x$ and $u \neq z$) and the subtree rooted at $u$ in the same way as in $r$. After returning to $y$ from $u$, traverse another adjacent node $v$ ($v \neq x$ and $v \neq z$) in the same fashion. After all adjacent nodes (other than $x$ and $z$) of $y$ have been exhausted, travel from $y$ to $z$ and continue along $l$. The open tour terminates when the other end of $l$ is reached. The open tour travels the same set of links as $r$ except that links on $l$ are traversed only once.

□

We illustrate the constructive proof using Figure 5. Suppose we are given the minimum length (18) closed tour $(C_5, C_2, C_6, C_7, C_8, C_7, C_6, C_2, C_1, C_9, C_{10}, C_9, C_1, C_3, C_1, C_4, C_1, C_2, C_5)$ and a terminal chain $(C_5, C_2, C_6, C_7, C_8)$ of length 4. By reconnecting the tour and traversing the chain only once, we obtain the open tour $(C_5, C_2, C_1, C_9, C_{10}, C_9, C_1, C_3, C_1, C_4, C_1, C_2, C_6, C_7, C_8)$. It has the length $18 - 4 = 14$.

Lemma 13 extends Lemma 12 to a minimum length open tour.

**Lemma 13** *An minimum length open tour of a tree has the length of a minimum length closed tour minus the length of a longest terminal chain.*

Proof: Assume an open tour is constructed as in the proof of Lemma 12, which is based on a terminal chain of the longest length of all terminal chains. It is sufficient to show that no single link traversal can be removed from this tour such that it remains to be an open tour.

Each link along the terminal chain is traversed only once. If any one of these link traversals is removed, the tour will be disconnected. Each link in a subtree other than the terminal chain is traversed twice, one of them travels away from the terminal chain, and the other travels towards the chain. If any of these link traversals is removed, the tour of the subtree will be disconnected. □

We now remove the assumption of identical link weight and the result for the tour problem follows.

**Theorem 14** *An minimum weight open tour of a weighted tree can be constructed by modifying a minimum weight closed tour such that a terminal chain of the maximum weight is traversed only once.*

Proof: This is a direct extension of Lemma 13 to weighted trees. The replacement of length by weight is valid because Lemma 10, 12, 13 and their proofs are still valid if the term *length* is replaced by the term *weight* and weights are different. □

In Figure 6, the terminal chain of the maximum weight $(1 + 2 + 4 + 8 + 6 = 21)$ is $(C_8, C_7, C_6, C_2, C_1, C_4)$. Therefore, a minimum weight open tour is $(C_8, C_7, C_6, C_2, C_5, C_2, C_1, C_3, C_1, C_9, C_{10}, C_9, C_1, C_4)$ and the minimum weight is $2 * (1+2+4+4+8+3+2+6+4) - 21 = 47$.

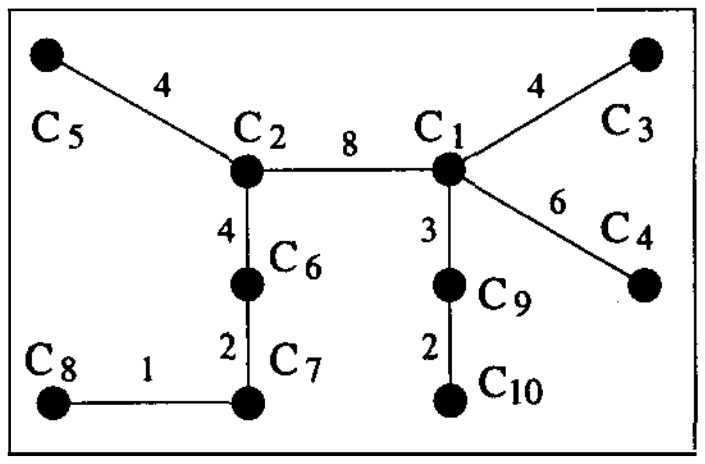

Figure 6: A weighted tree for the problem of the minimum weight tour.

Based on Theorem 14, Algorithm 15 finds a minimum weight open tour for a given tree. The steps 1 through 5 of the algorithm find a terminal chain with the maximum weight. It can be viewed as a variation of Dijkstra's shortest-path algorithm in a tree. It differs from the latter in that it finds the longest (heaviest) path between a non-predetermined pair of leaves. The step 6 constructs a minimum weight open tour in a way as described in the proof of Lemma 12. It also produces a numbering of nodes as stated in Problem Statement 7.

The algorithm has a complexity of $O(n^2)$ for both time and space.

Figure 7 illustrates Algorithm 15. The weighted tree is identical to that in Figure 6 up to a renaming of nodes. The renaming is performed such that the five leaves are indexed from 1 to 5, satisfying the input description of Algorithm 15. Note that the indexing of nodes in Figure 6 is consistent with the tree structure, but the indexing in Figure 7 is not.

Following Algorithm 15, we obtain $M[1..5] = (18, 21, 20, 21, 19)$ as shown in Figure 7. Therefore, $x = 2$, $y = 4$ and a heaviest terminal chain is one from $v_2$ to $v_4$. This is the same as we obtained earlier from Figure 6. The step 6 may (since the result is not unique) fill the array $t$ as $t[1..14] = (2, 8, 7, 6, 1, 6, 9, 5, 9, 10, 3, 10, 9, 4)$ which corresponds to the same minimum weight open tour as we obtained

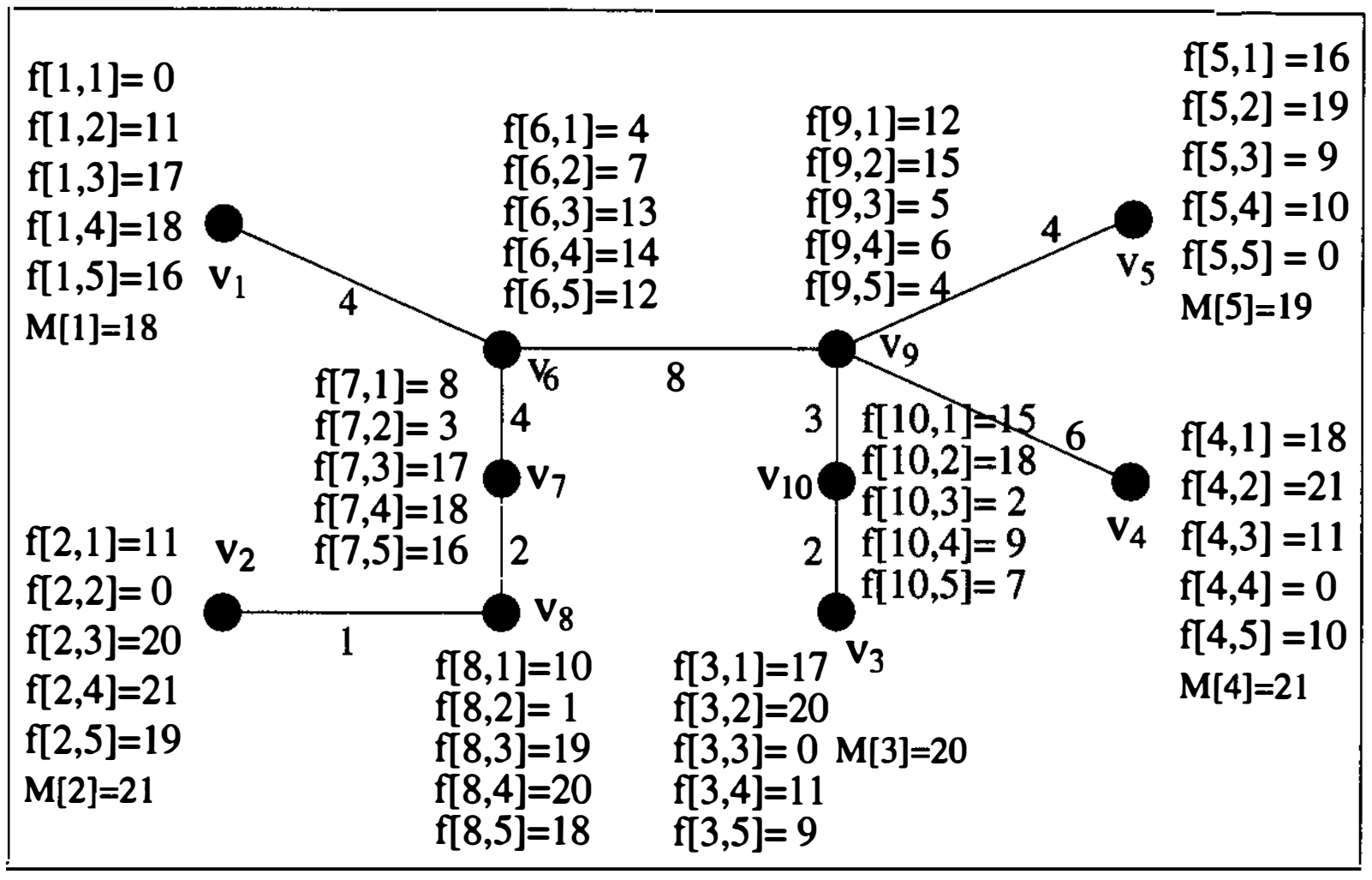

Figure 7: A weighted tree to illustrate Algorithm 16.

**Algorithm 15**

*Input:* *A weighted tree of a set $N$ of $n > 2$ nodes with $m$ leaves $v_1, \ldots, v_m$ and $n - m > 0$ internal nodes $v_{m+1}, \ldots, v_n$.*

*Var:* $f[1..n, 1..m]$, $M[1..m]$ : *array of reals;* $t[1..2n]$ : *array of integers.*

*Output:* *An open tour defined by t[] and a numbering of nodes.*

*begin*

*1 for $j = 1$ to $m$, do $f[j,j] := 0$*

*2 for $j = 1$ to $m$, do*
  *while there exists $i$ $(1 \leq i \leq m)$*
    *and $f[i,j]$ is undefined, do*
      *if $v_i$ is adjacent to a node $v_k$ and $f[k,j]$ is defined, then $f[i,j] := f[k,j] + w(v_i, v_k)$*

*3 for $i = 1$ to $m$, do $M[i] := \max_{j=1}^{m} f[i,j]$*

*4 Find the leaf $v_x$ such that $M[x] = \max_{i=1}^{m} M[i]$*

*5 for $j = 1$ to $m$, do*
  *if $f[x,j] = M[x]$ then $y := j$, break*

*6 Travel along the terminal chain from $v_x$ to $v_y$. At each internal node $z$ on the chain, traverse each subtree rooted at an adjacent node (not on the chain) of $z$ in a depth-first fashion. Record the index of a node in t[] each time it is visited. Number each node as it is visited the first time.*

*7 Return t[] as the open tour and the numbering.*

*end*

earlier from Figure 6. The numbering produced is then $(2, 8, 7, 6, 1, 9, 5, 10, 3, 4)$ which satisfies Problem Statement 7.

To use the open tour obtained for a new version of `UpdateBelief`, we must make sure that the order in which each node is visited *the first time* (the numbering of nodes returned by Algorithm 15) is consistent with the tree structure. This order corresponds to the order in which linkages are indexed and the order in which `AbsorbThroughLinkage` will be performed. This condition is required in the proof of Theorem 4. As we can see that it is indeed true since the next node to number (in step 6 of Algorithm 15) is always adjacent to the subtree traversed so far.

Theorem 16 summarizes the above discussion on Algorithm 15. The proof is trivial given Theorem 14, and the equivalence of Problem Statement 7 and 9.

**Theorem 16** *Let $T$ be a weighted tree. The numbering of nodes generated by Algorithm 15 is consistent with $T$, and the open tour returned has the minimum weight.*

We are now ready to define another improved version of `UpdateBelief`. Since belief propagation to the host tree is not necessary, we define a new operation `DistributeEvidenceOnChain`. It is the same as `DistributeEvidence` except we only propagate belief along a specified chain. We define the new `UpdateBelief` and state its correctness as follows.

**Operation 17 (UpdateBelief3)** *Let $\{L_1, \ldots, L_m\}$ be the set of linkages of a JT $T^a$ relative to $T^b$. Let $U_i^a$ and $U_i^b$ be the linkage hosts of $L_i$ $(i = 1, \ldots, m)$ in $T^a$ and $T^b$, respectively. When* `UpdateBelief3` *is initiated by $T^a$ relative to $T^b$, the following is performed.*

*For $i = 1$ through $m$,* `AbsorbThroughLinkage` *is called in $U_i^a$ to absorb from $U_i^b$ through $L_i$. For $i = 1, \ldots, m-1$, it is followed by* `DistributeEvidenceOnChain` *called in $U_i^a$ along the unique path from $U_i^a$ to $U_{i+1}^a$. For $i = m$, it is followed by* `DistributeEvidence` *called in $U_m^a$.*

**Corollary 18** *Let $I$ be the d-sepset between JTs $T^a$ and $T^b$. Let $\{L_1, \ldots, L_m\}$ be the set of linkages indexed according to the numbering produced by Algorithm 15. Let the two JTs be internally consistent. Let $B(I^a)$ $(B(I^b))$ be the belief table on $I$ defined by marginalization of $B(T^a)$ $(B(T^b))$.*

*After* `UpdateBelief3`*, we have $B'(T^a) = B(T^a) * B(I^b)/B(I^a)$, and $T^a$ is internally consistent.*

Given a JT and a d-sepset with an adjacent JT, `UpdateBelief3` is optimal in the sense that the min-imum amount of computation for coordinating multilinkage belief propagation is required.

## 7 Comparison of Versions of UpdateBelief

This paper addresses the computation efficiency of belief propagation between a pair of Bayesian subnets in a MSBN. Inter-subnet belief propagation involves the passage of the probability distribution on d-sepset $I$ from one subnet to an adjacent subnet.

A brute force method forms a large clique that contains $I$ in each subnet involved, and pass the belief table $B(I)$ directly. It is computationally the most expensive, both locally and inter-subnet-wise.

The first improvement is `UpdateBelief`. Only belief tables on linkages are passed, which are collectively smaller than $B(I)$. It reduces the inter-subnet traffic but still incurs expensive local computation to coordinate belief propagation over multiple linkages.

The second improvement is `UpdateBelief2`. Multiple performances of `DistributeEvidence` are replaced by multiple performances of the operation `DistributeEvidenceOnHostTree`. Distribution beyond the host tree is saved at each performance. The saving of inter-subnet traffic obtained by `UpdateBelief` is maintained while the efficiency of local computation is improved.

The next improvement is `UpdateBelief3`. Multiple performances of `DistributeEvidenceOnHostTree` are replaced by multiple performances of the operation `DistributeEvidenceOnChain`. Each time, distribution to the entire host tree is reduced to distribution along a chain leading to the next host. It further improves the efficiency of local computation beyond `UpdateBelief2`, and incurs the minimum amount of computation to coordinate belief propagation over multiple linkages.

Our analysis for the minimum weight open tour, which leads to the definition of `UpdateBelief3`, has assumed equal weights in traversing a link in both directions. The assumption may not hold since belief propagation between a pair of cliques may incur different amount of computations when performed in opposite directions. The results of this paper can be extended to cover the situation where weights differ at opposite directions. Due to the limited space, such discussion is beyond the scope of this paper.

We indicate that the savings in computation time are obtained by increased sophistication in control mechanisms. Replacement of the brute force method (equivalent to a single linkage propagation) by `UpdateBelief` requires the coordination of multiple linkage propagation. Replacement of `UpdateBelief` by `UpdateBelief2` requires additional control. Distribution is to be terminated at the leaves of the host tree. Finally, replacement of `UpdateBelief2` by `UpdateBelief3` requires more specific control. Distribution must proceed along predetermined chains.

### Acknowledgements

This work is supported by the Dean's Research Funding from Faculty of Science, University of Regina, the General NSERC Grant from University of Regina, and Research Grant OGP0155425 from NSERC. I am also grateful to the suggestions from anonymous referees.